\documentclass[12pt]{article}

\newcommand{\blind}{0}

\usepackage{amsthm,amsmath,amssymb,bbm}
\usepackage{setspace, fancybox, fullpage}
\usepackage{comment}
\usepackage{multirow}
\usepackage{rotating}
\usepackage{subfigure}
\usepackage{colortbl}

\newtheorem{theorem}{Theorem}

\newtheorem{proposition}{Proposition}

\def\xv{\boldsymbol x}

\def\Xv{\boldsymbol X}

\def\Am{\mathrm A}

\newcommand{\gammav}{\mbox{\boldmath{$\gamma$}}}

\newcommand{\muv}{\mbox{\boldmath{$\mu$}}}

\newcommand{\omegav}{\mbox{\boldmath{$\omega$}}}

\newcommand{\eps}{\varepsilon}

\def\1v{\mathbf 1}
\def\0v{\mathbf 0}

\newcommand{\R}{\mathbb R}

\newcommand{\E}{\mathbb E}
\newcommand{\sgn}{\mathop{\mathrm{sign}}}
\def\Pr{\mathrm P}

\newcommand{\wh}{\widehat}
\newcommand{\argmin}{\operatornamewithlimits{argmin}}

\def\half{\frac{1}{2}}

\def\spacingset#1{\renewcommand{\baselinestretch}%
{#1}\small\normalsize} \spacingset{1}

\begin{document}
\def\spacingset#1{\renewcommand{\baselinestretch}%
{#1}\small\normalsize} \spacingset{1}

\if0\blind
{
  \title{\bf Sparse Fisher's Linear Discriminant Analysis for Partially Labeled Data}
  \author{Qiyi Lu\thanks{Correspondence to: Qiyi Lu (e-mail: qlu@math.binghamton.edu). Qiyi Lu is a doctoral candidate and Xingye Qiao (e-mail: qiao@math.binghamton.edu) is an Assistant Professor at Department of Mathematical Sciences at Binghamton University, State University of New York, Binghamton, New York, 13902-6000. Qiao's research is partially supported by a collaboration grant from \textit{Simons Foundation} (award number 246649).}\quad and Xingye Qiao\\
    Department of Mathematical Sciences\\
    Binghamton University, State University of New York\\
    Binghamton, New York, 13902-6000\\
    \quad\\
    E-mails:\{qlu,qiao\}{@}math.binghamton.edu\\
    Phone: (607) 777-2147\\
    Fax: (607) 777-2450}
  \maketitle
  \newpage
} \fi

\if1\blind
{
  \bigskip
  \bigskip
  \bigskip
  \begin{center}
    {\LARGE\bf Sparse Fisher's Linear Discriminant Analysis for Partially Labeled Data}
  \end{center}
  \medskip
} \fi

\bigskip
\begin{abstract}
Classification is an important tool with many useful applications. Among the many classification methods, Fisher's Linear Discriminant Analysis (LDA) is a traditional model-based approach which makes use of the covariance information. However, in the high-dimensional, low-sample size setting, LDA cannot be directly deployed because the sample covariance is not invertible. While there are modern methods designed to deal with high-dimensional data, they may not fully use the covariance information as LDA does. Hence in some situations, it is still desirable to use a model-based method such as LDA for classification. This article exploits the potential of LDA in more complicated data settings. In many real applications, it is costly to manually place labels on observations; hence it is often that only a small portion of labeled data is available while a large number of observations are left without a label. It is a great challenge to obtain good classification performance through the labeled data alone, especially when the dimension is greater than the size of the labeled data. In order to overcome this issue, we propose a semi-supervised sparse LDA classifier to take advantage of the seemingly useless unlabeled data. They provide additional information which helps to boost the classification performance in some situations. A direct estimation method is used to reconstruct LDA and achieve the sparsity; meanwhile we employ the difference-convex algorithm to handle the non-convex loss function associated with the unlabeled data. Theoretical properties of the proposed classifier are studied. Our simulated examples help to understand when and how the information extracted from the unlabeled data can be useful. A real data example further illustrates the usefulness of the proposed method.
\end{abstract}


\vfill
\newpage

\spacingset{1.87} 

\setcounter{page}{1}
\abovedisplayskip=8pt
\belowdisplayskip=8pt

\section{Introduction}
Classification is an important tool in modern statistical analysis. In a classification problem, the training data set $\{(\xv_i,y_i),~i=1,\ldots,n\}$ is obtained from an unknown distribution, where $\xv_i\in\R^d$ is the observed covariates and $y_i$ is the class label for the $i$th observation. In this article, we focus on binary classification, that is a classification problem with only two possible classes, $y_i\in\{+1,-1\}$.  The goal of classification is to obtain a classification rule $\phi(\cdot)$ based on the training data, such that for any new observation with only the convariates $\xv$ available, its class label $y$ can be accurately predicted as $\phi(\xv)$.

There are many classification methods in the literature. For an overall introduction, see \cite{Hastie2009elements}. One popular group of methods are the linear classifiers, due to their simplicity and interpretability. For a linear classifier, the classification rule is defined as $\sgn\{f(\xv)\}$, where $f(\xv) = \omegav'\xv+b$, $\omegav\in \R^d$ and $b\in \R$, is a discriminant function linear in $\xv$, obtained from the training data. Some examples of linear classifiers include Fisher's Linear Discriminant Analysis (LDA) \cite{fisher1936use}, Logistic Regression \cite{berkson1944application}, Support Vector Machine (SVM) \cite{vapnik1995nature,Cortes1995Support}, $\psi$-learning \cite{shen2003on}, Distance Weighted Discrimination (DWD) \cite{marron2007distance}, Large-margin Unified Machine \cite{liu2011hard}, and hybrids of SVM and DWD \cite{Qiao2015Flexible,Qiao2015Distance}. For a binary linear classifier, a classification boundary (also known as a separating hyperplane) is induced by $\{\xv:\omegav'\xv+b=0\}$ which divides the sample space $\R^d$ into two halves, one for each class.

Despite the new and fast development of the latter methods above, the LDA method is still widely used among practitioners. LDA is a traditional model-based classification approach which makes use of the covariance information under a Gaussian assumption. Because LDA is simple to implement and straightforward to interpret, it is one of the most popular statistical methods for classification. Lee and Wang \cite{lee2015does} compared the performance of LDA with some machine learning approaches and concluded that LDA would work better in cases where the Gaussian assumption is roughly true. That being said, LDA has several drawbacks which makes it undesirable in more complicated data settings (see the next two subsections). The goal of this article is to enrich the potential of LDA in these settings.

\subsection{Working with High-Dimensional, Low-Sample Size Data}
The High-Dimensional, Low-Sample Size (HDLSS) data setting is very challenging for statistical learning and it appears in many applied fields such as gene expression micro-array analysis, facial recognition, medical image analysis and text mining. In the HDLSS context, classical multivariate statistical methods often fail to give a meaningful analysis \cite{marron2007distance}. For example, there exists an interesting phenomenon called `data piling' for discriminant analysis \cite{ahn2010maximal}. `Data piling' means that when training data points are projected onto a low-dimensional discriminant subspace, many of the projections are identical. This phenomenon is caused by the fact that the corresponding discriminant subspace, a one-dimensional coefficient vector in the case of binary classification, is driven by very particular artifacts of the realization of the training data. This makes `data piling' an undesirable property for discrimination since the classifier performs worse for out-of-sample test data. Other challenges in the HDLSS setting include the collinearity among predictors, the error aggregation over dimensions, among others (see \cite{fan2010selective}). Fan and Li \cite{fan2006statistical} gave a comprehensive overview of statistical challenges with high dimensionality in diverse disciplines, such as computational biology, health studies and financial engineering. In particular, they demonstrated that for many statistical problems, the model parameters can be estimated as if the best model was known in advance, as long as the dimensionality was not excessively high. These challenges have motivated the development of new methods and theory in the HDLSS setting.

In recent years, many efforts have been made to make classification methods more suitable for the HDLSS data. The DWD method \cite{marron2007distance} claimed to enjoy a better discriminant subspace than SVM. In addition, a great number of research articles are dedicated to improving traditional methods so that they have sparse discriminant coefficient vectors. An underlying assumption is that there are only a few variables which truly drive the difference between classes. Hence, a variety of methods use regularization approaches to encourage a sparse representation of the coefficient vector $\omegav$, which in general are obtained from optimizations of the form,
$$\argmin_{\omegav,b} \sum_{i=1}^n L(\omegav,b,\xv_i,y_i)+\lambda \cdot p(\omegav),$$
where $L(\cdot)$ is a loss function to minimize the misclassification and $p(\cdot)$ is a penalty term to control the model complexity. Common choices of the penalty function include the $\ell_1$ norm penalty, the SCAD penalty \cite{fan2001variable} and the minimax concave penalty (MCP) \cite{zhang2010nearly}. Examples of sparse regularization methods include the lasso \cite{tibshirani1996regression} and the elastic net \cite{zou2005regularization} in regression, and the $\ell_1$ norm SVM \cite{zhu20041}, the sup-norm SVM \cite{zhang2008variable} and the direct sparse discriminant analysis (DSDA) \cite{mai2012dsda} in classification. See \cite{Qiao2014Variable} for a review.

\subsection{Working with Partially Labeled Data}
In many real problems, it is difficult or expensive to obtain the class label information; on the other hand, it may be relatively cheap to obtain the covariate information quickly for many observations. Hence, it is often the case that there are many observations without labels (unlabeled data) and a few observations with labels (labeled data). For instance, in spam detection, there are a large number of unidentified emails, but only a small set of identified emails are used to train a filter to flag incoming spams. In facial recognition, the training data may include a few faces with scars identified manually and enormous unidentified faces. In these situations, one typical research question is how to use both unlabeled and labeled data to enhance the prediction accuracy.

In the big data era, the dimension of the data is often greater than the sample size of the labeled data, though not necessarily greater than that of the unlabeled data. In this case, we have an HDLSS setting, if considering the labeled data only. On the other hand, the unlabeled data may contain useful information to overcome the difficulty caused by the high dimensionality. Semi-supervised learning is a class of machine learning techniques that make use of both labeled and unlabeled data for modeling. This article is written with the belief that unlabeled data, in some cases, can indeed produce considerable improvement in learning accuracy. Note that the semi-supervised learning problem is different from the more traditional missing data problem in statistics: in the current article, the size of unlabeled data is much greater than that of the labeled data.

Many semi-supervised approaches have been proposed in different settings, including the co-training method \cite{blum1998combining}, the  EM algorithm \cite{nigam2000text}, the bootstrap method \cite{collins1999unsupervised}, the Bayesian network \cite{cozman2003semi}, the Gaussian random field and harmonic function \cite{zhu2003semi}, the transductive SVM (TSVM) \cite{vapnik1998statistical, chapelle2006semi, wang2007transductive}, the large-margin based methods \cite{wang2007large, wang2009efficient} and the graph-based regularization methods \cite{chapelle2003cluster,belkin2006manifold,cai2007semi,sugiyama2010semi}. Many of these methods rely on the clustering assumption \cite{chapelle2004semi} which assumes the closeness between the classification and the grouping (clustering) boundaries.

In this article, we aim to improve a model-based classifier, namely the classical LDA method, so that it can be used to classify partially labeled data in a high-dimensional space. This is achieved by marrying LDA with a machine learning technique to incorporate the unlabeled data. The end product of the article is a Semi-Supervised Sparse Linear Discriminant Analysis ($S^3$LDA) method. 

The rest of the article is organized as follows. Section 2 starts with an introduction to the existing development for sparse LDA, followed by our proposed $S^3$LDA method. Section 3 presents the implementation of our method, including the tuning parameter selection issue. Some theoretical results are presented in Section 4, followed by numerical studies in Section 5. Section 6 contains some concluding remarks. The Appendix is devoted to technical proofs.

\section{Semi-supervised LDA in HDLSS Setting}
Consider a binary classification problem.  Let $\Xv=(X_1,\cdots,X_d)'\in\R^d$ be the covariates, $Y\in\{+1,-1\}$ be the class label and $n_+$ and $n_-$ be the sizes of the positive and negative classes. The LDA method assumes that $\Xv|Y=y\sim N(\muv_y,\Sigma)$, $\Pr(Y=+1)=\pi_1$, and $\Pr(Y=-1)=\pi_2=1-\pi_1$. Given $\Sigma$, $\muv_+$ and $\muv_-$, the Bayes classification rule is given by $\phi^{\textrm{Bayes}}(\xv) = \sgn({\omegav^{\textrm{Bayes}}}'\xv+b^{\textrm{Bayes}})$, where the coefficient vector $\omegav^{\textrm{Bayes}}=\Sigma^{-1}(\muv_+-\muv_-)$ and the intercept term $b^{\textrm{Bayes}}=-(\muv_++\muv_-)'\Sigma^{-1}(\muv_+-\muv_-)/2+\log (\pi_1/\pi_2)$. Hence the Bayes rule classifies an observation $\xv$ to the positive class if and only if $\left\{\xv-(\muv_++\muv_-)/2\right\}'\Sigma^{-1}(\muv_+-\muv_-)+\log (\pi_1/\pi_2)>0.$ In practice, since the true distributions are unknown, we use the pooled sample covariance $\wh{\Sigma}$, the sample mean vectors $\hat{\muv}_+$ and $\hat{\muv}_-$, and $n_+/n_-$ to estimate $\Sigma$, $\muv_+$, $\muv_-$, and $\pi_1/\pi_2$, respectively. The resulting classification rule is the LDA classifier.


For many data sets in modern applications, the dimension $d$ can be much greater than the sample size $n$. In such cases, LDA cannot be directly used because the sample covariance $\wh{\Sigma}$ is not invertible with probability one. Moreover, the sample mean difference $(\hat{\muv}_+-\hat{\muv}_-)$ may be deviated from the true population mean difference at each dimension, which could lead to overfitting of the classifier due to error aggregated over the $d$ dimensions.

\subsection{Sparse LDA}
It is a common practice to overhaul a traditional statistical method by introducing sparsity for high-dimensional data. The pioneers of sparse LDA include the nearest shrunken centroids classifier \cite{tibshirani2002diagnosis}, the `naive Bayes' classifier \cite{bickel2004some}, and the features annealed independent rule (FAIR) \cite{fan2008fair}. These methods are based on the independence rule that ignores the correlation among features. The nearest shrunken centroids classifier and the `naive Bayes' classifier use only the diagonal of the sample covariance to estimate $\Sigma$ while the FAIR method conducts feature selection based on the marginal $t$-statistics in two-sample $t$ tests. Although these classifiers are easy to interpret and computationally attractive, they may lose critical covariance information and hence may be suboptimal. Strong correlations can exist in high-dimensional data and ignoring them may lead to misleading feature selection. In particular, Mai and Zou \cite{mai2012dsda} pointed out that variable selection can be inconsistent when the correlation is ignored; moreover, as the sample size goes to infinity, the Bayes risk may not be achieved in this case. 

Recent years, many efforts have been devoted to developing sparse versions of LDA, such as the $\ell_1$-Fisher's discriminant analysis (FSDA) \cite{wu2009sparse}, the regularized optimal affine discriminant (ROAD) \cite{fan2010road}, the penalized LDA methods \cite{witten2011penalized}, the direct sparse discriminant analysis (DSDA) \cite{mai2012dsda} and the sparse optimal scoring (SOS) \cite{clemmensen2011sparse}. Moreover, Mai and Zou \cite{mai2013note} revealed the connection and equivalence between FSDA, DSDA and SOS. 

Although sparse LDA has been a popular research topic, to our best knowledge, little progress has been made to generalize the LDA method to a scenario with many unlabeled data, a gap which the current article intends to fill.

\subsection{Proposed Method}

Consider a binary classification problem with the labeled data $\{(\xv_i,y_i),~i=1,...n_l\}$, and the unlabeled data $\{\xv_{n_l+j},~ j=1,\cdots,n_u\}$. The total sample size is  $n=n_l+n_u$. Our goal is to find a linear classification function to classify a partially labeled dataset, which is of the form $f(\xv)=\omegav'\xv+b$, by solving the following optimization problem, 
\begin{equation}
\min_{\omegav,b}~C_1\sum_{i=1}^{n_l}L(y_i,f(\xv_i))+C_2\sum_{i=n_l+1}^{n_l+n_u} U(f(\xv_i))+p(\omegav,b),
\label{alg0}
\end{equation}
where $L(\cdot)$ is a loss function for the labeled data to control the misclassification, and $U(\cdot)$ is a loss function for the unlabeled data to encourage large margin between two clusters. As usual, $p(\cdot)$ is a penalty term which controls the model complexity. The non-negative tuning parameter set $C=(C_1, C_2)$ balances the trade-off among the misclassification, the large margin between clusters and the model complexity.

It is a general framework to obtain the linear classification function from \eqref{alg0}, where many different loss functions and penalty functions may be used for $L(\cdot)$, $U(\cdot)$ and $p(\cdot)$. Examples of $L(\cdot)$ include, among others, the logistic loss $L(y,f)=\log(1+e^{-yf})$ \cite{zhu2005kernel}; the hinge loss $L(y,f)=(1-yf)_+$ for SVM with its variants $L(y,f)=(1-yf)_+^q$ for $q>1$ \cite{lin2002support}; the $\psi$-loss $L(y,f)=1-\sgn(yf)$ when $yf\geq1$ or $yf<0$, and $2(1-yf)$ otherwise \cite{shen2003on}.

Though the squared error loss defined as $L(y,f)=(\tilde y-f)^2$ (for $\tilde{y}=n/n_+$ when $y=+1$, and $-n/n_-$ when $y=-1$) is more widely used in the regression setting, it can also be used for classification. This is because the classical LDA method can be exactly reconstructed via the least squares regression (see Chapter 4 of \cite{Hastie2009elements}). In particular, let $\omegav^{\textit{OLS}}$ be the coefficient vector obtained from the regression problem
$$\left(\omegav^{\textit{OLS}},b^{\textit{OLS}}\right)=\argmin_{\omegav,b} \sum_{i=1}^n (\tilde{y}_i-b-\omegav'\xv_i)^2,$$
where $\tilde{y_i}=n/n_+$ when $y_i=+1$, and $-n/n_-$ when $y_i=-1$. It can be verified that $\omegav^{\textit{OLS}}=c\hat{\Sigma}^{-1}(\hat{\muv}_+-\hat{\muv}_-)$ for some positive constant $c$, which is along the same direction as the LDA coefficient vector $\hat{\Sigma}^{-1}(\hat{\muv}_+-\hat{\muv}_-)$. In our proposed $S^3$LDA approach, we choose to use the squared error loss, after coding $y\in\{+1,-1\}$ as $\tilde y\in\{n/n_+,-n/n_-\}$. The same loss was previously considered by DSDA \cite{mai2012dsda} for classification.


The second term in (\ref{alg0}), involving $U(\cdot)$, is associated with the unlabeled data, and is included to encourage a large margin between two clusters induced by the classification rule. This is done by assigning a large loss when the classification boundary goes through an area with high density, hence, encouraging the classification boundary to avoid those areas and go through a gap between two clusters. In particular, $U(z)$ is a function with maximal value at $z=0$ and a decreasing value as $|z|$ increases. In order to have this property, we can modify existing loss functions for classification, such as the hinge loss, the logistic loss and the $\psi$-loss, by changing $yf(\xv)$ to $|f(\xv)|$ in their definitions. For example, the modified logistic loss is $U(z)=\log(1+e^{-|z|})$ and the modified hinge loss is $U(z)=(1-|z|)_+$. That is, we assign a zero loss when $|z|>1$ and a loss of $1-|z|$ otherwise. Wang and Shen \cite{wang2007large} and Wang et al. \cite{wang2009efficient} also considered the modified hinge loss in their machine learning-oriented methods for large-margin semi-supervised learning.

Although $U(\cdot)$ is applied to the unlabeled data only, we expect it to improve the classification boundary by making the margin wider. For illustration purpose, in the rest of the article, we use the modified hinge loss for $U(\cdot)$. But other choices for $U(\cdot)$ are possible.

%

In summary, in our proposed $S^3$LDA method, we combine a classical model-based approach, LDA, and a machine learning-oriented method, to classify high-dimensional partially labeled data. They are reflected in the choice of the loss functions $L$ and $U$ in \eqref{alg0}.  To be specific, we choose $L(y,f(\xv))=(\tilde y-f(\xv))^2$ and $U(f(\xv))=(1-|f(\xv)|)_+$. 

\subsection{Penalty Term}
Our penalty term $p(\omegav,b)$ is chosen as $\|\omegav\|_1+c\|\hat\omegav\|^{-1}|b|$. The first term herein is an $\ell_1$ norm penalty of $\omegav$, to find variables on which the two classes are significantly different and shrink the coefficients for the other variables to zero. Other functions than the $\ell_1$ norm, such as the elastic net, SCAD and MCP penalties, may be used as well. 

The second term in the penalty function is included to prevent an undesirable case as follows. Note that when the parameters are not chosen wisely (such as when $C_1=0$), a problem may occur that the classification boundary is pushed to be infinitely far away from the data set, because this would induce a zero loss in $U(\cdot)$, with no cost on $L(\cdot)$. However, it is obvious that the classification performance is poor in this case since one class is totally ignored. To avoid this issue, without loss of generality, we assume that each predictor has mean 0 and variance 1; we include an adaptive penalty on the intercept term $b$, that is, $\|\hat\omegav\|^{-1}|b|$, as the second term of $p(\omegav,b)$, to encourage a small value for $b$. Here $\hat\omegav$ is an initial estimate of the classification coefficient vector $\omegav$. Note that, $b=0$ indicates that the classification boundary goes through the origin point, in which case the undesirable situation is avoided.

\section{Implementation}
In this section, we discuss the implementation of our method. We first introduce the algorithm for optimizing the unusual objective function in (\ref{alg0}). We then discuss the problem of  tuning parameter selection.
\subsection{Algorithm}
Solving the problem in $\eqref{alg0}$ with the $U$ loss being the modified hinge loss involves a non-convex optimization. To overcome this difficulty, we make use of the difference of convex functions (DC) algorithm \cite{tao2005dc}. The key to the DC algorithm is to decompose the non-convex objective function to the difference of two convex functions which leads to a sequence of convex optimizations. The sequence of local solutions converges to a stationary point. The DC algorithm has been used in several other works to solve non-convex optimization problems, such as \cite{wang2007large} and \cite{wu2007robust}. For the sake of self-containment, we provide the brief idea of the DC algorithm when applied to our method.

\begin{figure}[!ht]\vspace{-1.5ex}\footnotesize
  \begin{center}
  \includegraphics[height=0.4\linewidth]{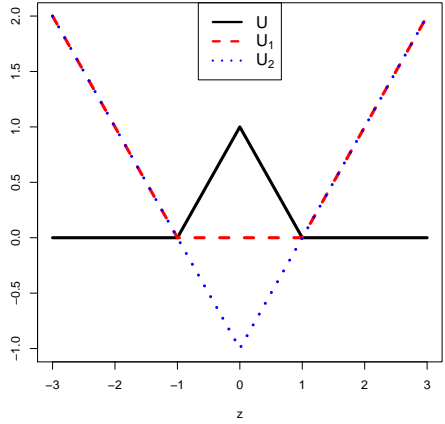}
  \end{center}
\vspace{-5ex}
\caption{The DC decomposition of $U=U_1-U_2$. Functions $U$, $U_1$ and $U_2$ are represented by the solid, dashed and dotted lines respectively.}
\label{DC}
\end{figure}

We first decompose the modified hinge loss $U=(1-|z|)_+$ into the difference between two convex functions $U_1-U_2$, where $U_1=(|z|-1)_+$ and $U_2=|z|-1$. The decomposition is displayed in Figure ~\ref{DC}.

Let $f(\xv)=\omegav'\xv+b$. Rewrite the objective function as 
$$Q=C_1\sum_{i=1}^{n_l}L(y_i,f(\xv_i))+C_2\sum_{i=n_l+1}^{n_l+n_u} U(f(\xv_i))+\|\omegav\|_1+c\|\hat\omegav\|^{-1}b,$$which can be decomposed similarly as $Q_1-Q_2$ where
\begin{align*}
Q_1&=C_1\sum_{i=1}^{n_l}L(y_i,f(\xv_i))+C_2\sum_{j=1}^{n_u} U_1(f(\xv_i))+\|\omegav\|_1+c\|\hat\omegav\|^{-1}b, \\
\makebox{where~}Q_2&=C_2\sum_{j=1}^{n_u} U_2(f(\xv_i)).
\end{align*}Note that both $Q_1$ and $Q_2$ are convex. However, to minimize the non-convex $Q_1-Q_2$, we use a linear approximation to $Q_2$, so that the approximated optimization problem is convex. Overall, the algorithm is conducted in a three-step iteration as follows.

\textbf{Step 1.} Set the initial values $(\omegav_0,b_0)$ of $(\omegav,b)$ to be the solutions of the sparse LDA with labeled data alone. Set a precision tolerance level $\eps>0$.

\textbf{Step 2.} At the $(k+1)$st iteration, compute $\left(\omegav^{(k+1)},b^{(k+1)}\right)$ by solving the convex problem
\begin{align*}
\left(\omegav^{(k+1)},b^{(k+1)}\right)=\argmin_{\omegav,b}~ Q_1{(\omegav,b;\omegav^{(k)})}-\left[\nabla Q_2|_{(\omegav^{(k)},b^{(k)})}\right]^T\cdot{\omegav\choose b},
\end{align*}
where $ Q_1{(\omegav,b;\omegav^{(k)})}$ is the result of substituting $\hat\omegav$ in $Q_1$ by $\omegav^{(k)}$ from the previous iteration and 
$\nabla Q_2|_{(\omegav^{(k)},b^{(k)})}$ is the gradient vector of $Q_2$ with respect to $(\omegav',b)'$, evaluated at the solution  $(\omegav^{(k)},b^{(k)})$ from the previous iteration.

\textbf{Step 3.} Repeat Step 2 until $|Q(\omegav^{(k+1)},b^{(k+1)})-Q(\omegav^{(k)},b^{(k)})|\leq \eps$.

It was shown \cite{wang2007large} that the number of iterations required to achieve the precision $\eps$ is $o(\log(1/\eps))$.

\subsection{Tuning Parameter Selection}
The $S^3$LDA method has three tuning parameters $C_1$, $C_2$ and $c$. The duo $(C_1,C_2)$ are our main focus, which jointly control the balance among the $L$ loss, the $U$ loss and the penalty. A common practice in the literature is to conduct a grid search on a set of parameter candidate values and compare their performance for an independent tuning data set. We search $(C_1,C_2)$ over $\mathcal{C}_1\times \mathcal{C}_2$ where  $\mathcal{C}_1=\{2^{-3},2^{-2},\dots,2^3\}$ and $\mathcal{C}_2=\{0,10^{-2},10^0,10^2\}$. We include zero in $\mathcal{C}_2$ so that our method encompasses a sparse LDA method (DSDA) \cite{mai2012dsda} which uses the labeled data alone.

Our penalty term is $|\omega_1|+\dots+|\omega_d|+c\|\hat\omegav\|^{-1}|b|$ with an ancillary parameter $c$. Note that $\|\hat\omegav\|^{-1}|b|$ is approximately the distance from the classification boundary to the origin point. For the examples in the numerical study, with each variable being normalized to have mean 0 and variance 1, we fix a universal value for $c=5$ which corresponds to a constraint that the distance is less than a universal fixed value. This choice is only reasonable when the data are normalized to have similar scales, which is the case in all our numerical studies.

Note that the tuning set is also partially labeled and the number of labeled data is very limited. If we ignore the unlabeled data in the tuning set and compare the misclassification rate for the labeled data only, the criterion may not be able to reflect the true goodness of the classifier. The choice of criterion can be critical for tuning parameter selection in some nontraditional situations, such as in imbalanced data classification problems \cite{qiao2009Adaptive, qiao2010weighted}. For an improvement in the partially labeled data, we propose to use a new criterion which involves two components where one component is the number of misclassified observations among the labeled data and the other one is a clustering measure for both the labeled and unlabeled data.

In particular, for each pair of $(C_1, C_2)$, we train a discriminant function, $f$. The number of misclassified labeled data points can be easily counted. The clustering measure is defined as the total number of tuning data points $\xv_i^{\textrm{tune}}$, whether labeled or unlabeled, which fall into a margin centered at the classification boundary with half-width $\eta$, that is we count the number of $|f(\xv_i^{\textrm{tune}})|<\eta$. Here $\eta$ is a typical measure of the scale, which is defined as one quarter of the sum of the $25$th and $75$th percentiles of the pairwise distance $|f(\xv_i^{\textrm{tune}})-f(\xv_{i'}^{\textrm{tune}})|$ for all $i\neq i'$. A small value of the clustering measure indicates that very few data points are close to the classification boundary and hence the margin induced by $f$ is indeed wide.

The choice of $\eta$ is a critical issue. Our choice of $\eta$ is adaptive to the underlying distribution of the data. It ensures that a reasonable portion of the data have a fair chance to fall in the gap which helps to identify the optimal tuning parameter pair.


\section{Theoretical Property}
In this section, we provide several theoretical justifications of the $S^3$LDA method. The classical LDA method is based on the Gaussian assumption. In this setting, without loss of generality, we assume that the two classes have means opposite to each other with respect to the origin. We further assume that the prior probabilities are the same. In this case, the Bayes classification boundary passes through the origin since $f_{\textrm{Bayes}}(\xv) = {\omegav^{\textrm{Bayes}}}'\xv+b^{\textrm{Bayes}}$ with $b^{\textrm{Bayes}}=0$. We have the following two propositions that describe the theoretical minimizers of risk functions with respect to the $L$ loss and the $U$ loss.

\begin{proposition}
Assume that $\Xv|Y=+\delta\sim \textsl{N}(\muv,\Sigma)$, $\Xv|Y=-\delta\sim \textsl{N}(-\muv,\Sigma)$ with $\Sigma$ full rank and $\Pr(Y=+\delta)=\Pr(Y=-\delta)=1/2$. Let $$\omegav_1=\argmin_{\omegav}\E(Y-\omegav'\Xv)^2$$ be the theoretical minimizer of the risk function with respect to the squared error loss, where the expectation is taken with respect to the distribution of $(\Xv,Y)$. Then $\omegav_1\propto \Sigma^{-1}\muv$.
\label{lemma1}
\end{proposition}                  

\begin{proposition}\label{lemma2}
Under the same assumption as Proposition \ref{lemma1}. Let $\omegav_2$ be the theoretical minimizer of the risk function with respect to the modified hinge loss, with the linear normalization constraint that $\omegav'\muv=1$,
\begin{align*}
	\omegav_2=\argmin_{\omegav:~\omegav'\muv=1}&\quad\E(1-|\omegav'\Xv|)_+,
\end{align*}where the expectation is taken with respect to the marginal distribution of $\Xv$.
Then $\omegav_2\propto \Sigma^{-1}\muv$.
\end{proposition}                  
               
Proposition~\ref{lemma1} and Proposition~\ref{lemma2} show that both theoretical minimizers of the risk functions for the $L$ loss and the $U$ loss in our $S^3$LDA method have the same coefficient vectors as that of the Bayes coefficient vector, up to some multiplicative constants. Theorem \ref{theorem1} shows that the theoretical coefficient vector of the unpenalized population version of our $S^3$LDA classifier is along the same direction as the Bayes coefficient vector.

\begin{theorem}
Assume that $\Xv|Y=+\delta\sim \textsl{N}(\muv,\Sigma)$, $\Xv|Y=-\delta\sim \textsl{N}(-\muv,\Sigma)$, where $\delta=1/(\muv'\Sigma^{-1}\muv)$, and $\Pr(Y=+\delta)=\Pr(Y=-\delta)=1/2$. Let $\omegav_\infty$ be the solution to the theoretical coefficient vector of the unpenalized population version of the $S^3$LDA classifier, defined by
\begin{align}\label{eq:theo_min}
	\omegav_\infty =\argmin_{\omegav:~\omegav'\muv=1}&\quad\E(Y-\omegav'\Xv)^2+C\E(1-|\omegav'\Xv|)_+,
\end{align}
where $C>0$ is a constant. Then $\omegav_\infty=\delta\Sigma^{-1}\muv$.
\label{theorem1}
\end{theorem}

To gain better insight, we reformulate the $\ell_1$ penalized $S^3$LDA classifier. Theorem \ref{theorem2} further reveals the small difference between the $\ell_1$ penalized version of the $S^3$LDA discriminant direction vector and the Bayes optimal direction vector.

\begin{theorem}
Let $s$ be the size of the set $K:=\{k:(\Sigma^{-1}\muv)_k\neq 0\}$ and $\lambda_{\max}(\Am)$ and $\lambda_{\min}(\Am)$ be the greatest and the smallest eigenvalues of matrix $\Am$. Assume that $\Xv|Y=+\delta\sim \textsl{N}(\muv,\Sigma)$, $\Xv|Y=-\delta\sim \textsl{N}(-\muv,\Sigma)$, where $\delta=1/(\muv'\Sigma^{-1}\muv)$, and $\Pr(Y=+\delta)=\Pr(Y=-\delta)=1/2$. Let $\omegav_\infty$ be as in (\ref{eq:theo_min}) and $\omegav^{\lambda}$ correspond to the $\ell_1$ penalized version, 
\begin{align}\label{eq:l1_theo_min}
	\omegav^{\lambda} =\argmin_{\omegav:~\omegav'\muv=1}&\quad\E(Y-\omegav'\Xv)^2+C\E(1-|\omegav'\Xv|)_++\lambda\|\omegav\|_1,
\end{align}
where $C>0$ is the same constant in (\ref{eq:theo_min}).  Then 
$$\|\omegav^{\lambda}-\omegav_\infty\|_2\le \frac{\lambda\sqrt{s}+C\sqrt{\lambda_{\max}(\tilde\Sigma)}}{\lambda_{\min}(\tilde\Sigma)},$$
where $\tilde\Sigma=\Sigma+\muv\muv'$.
\label{theorem2}
\end{theorem}

Theorem \ref{theorem2} characterizes the difference between the $S^3$LDA solution and the Bayes rule. This difference is jointly controled by parameters $\lambda$ and $C$. When $C=0$, the problem is reduced to the $\ell_1$-LDA when applied to the labeled data only, in which case $\|\omegav^{\lambda}-\omegav_\infty\|_2\rightarrow 0$ as $\lambda\rightarrow 0$. Similar results can be seen in the ROAD classifier \cite{fan2010road}.

\section{Numerical Study}
In this section we use simulated and real data examples to demonstrate the effectiveness of our proposed method. We compare it with linear $\ell_1$-SVM and a sparse LDA method (DSDA) \cite{mai2012dsda}, both of which are applied to the labeled data only, and a semi-supervised method, SELF \cite{sugiyama2010semi}. As a benchmark comparison, we also apply these methods to the full data set with all the labels available in order to see how much room our $S^3$LDA can improve. The misclassification rate, averaged over 100 replications, is reported.
\subsection{Simulations}
We consider four simulated examples. In each example, we first generate an i.i.d. random sample $\left\{(\xv_i,y_i),~i=1,\dots,n^*\right\}$ where $\xv_i=(x_{i1},\dots,x_{id})'$ and $d$ and $n^*$ vary, depending on the setting. Then we partition the data into training, tuning and test data sets. Importantly, we intentionally remove some class labels in the training and tuning sets to create a partially labeled data setting.

In Example 1 and Example 2, 3400 independent instances are generated. We use 200 for training, 200 for tuning and the remaining 3000 for testing.

\textbf{Example 1: } $(Y_i+1)/2\sim \textsl{Bernoulli}(0.5)$, $X_{i1}\sim \textsl{N}(1.4Y_i,1)$, and $X_{i2}\sim \textsl{N}(0,1)$, $i=1,\dots,3400$. Among the 200 training instances, 190 unlabeled instances $(X_{i1},X_{i2})$ are obtained by removing labels from a randomly chosen subset of the training sample, whereas the remaining 10 instances are treated as the labeled data. The 200 tuning instances are processed in the same way as the training set. 

\textbf{Example 2: } $(Y_i+1)/2\sim \textsl{Bernoulli}(0.5)$, $X_{i1}\sim \textsl{N}(sY_i,1)$, $X_{i2}\sim \textsl{N}(-sY_i,1)$ and $X_{i3},\dots,X_{i100}\sim \textsl{N}(0,1)$, $i=1,\dots,3400$. 
Here $s$ indicates the signal level and varies from 1 to 2.5. Among the 200 training instances, 190 unlabeled instances are obtained by removing labels from a randomly chosen subset of the training sample, whereas the remaining 10 instances are treated as the labeled data. The 200 tuning instances are processed in the same way as the training set.

In Example 3 and Example 4, We generate 10000 instances and study the performance of the methods with the increase of the dimensionality. In particular, we simulate datasets with dimension $d=20,30,40,50,100,200,500$. The number of training instances equals $200,244,283,316,447,632,1000$ respectively, which increases at the rate of $\sqrt{d}$. The tuning sample size is the same as the training sample size in each case, and the remaining instances are used for testing.

\textbf{Example 3: } $(Y_i+1)/2\sim \textsl{Bernoulli}(0.5)$, $X_{i1},X_{i3},X_{i5},X_{i7},X_{i9} \sim \textsl{N}(Y_i/2.7,1)$, and $X_{i2},X_{i4},X_{i6},$ $X_{i8},X_{i10}\sim \textsl{N}(-Y_i/2.7,1)$. The covariance of $X_{i1},\dots,X_{i10}$ is the Toeplitz matrix with the first row $(1,0.8,0.8^2,\dots,0.8^9)$. $X_{i11},\dots,X_{i100}\stackrel{\text{\tiny i.i.d}}{\sim}  \textsl{N}(0,1)$, $i=1,\dots,10000$. We randomly select 10 labeled data from each class and the remaining are treated as unlabeled data for both training and tuning sets.

\textbf{Example 4: } $(Y_i+1)/2\sim \textsl{Bernoulli}(0.5)$, $X_{i1},X_{i3},X_{i5},X_{i7},X_{i9} \sim  \sqrt{3/5}\times\textsl{t}(5)+Y_i/2.7$, and $X_{i2},X_{i4},X_{i6},$ $X_{i8},X_{i10}\sim \sqrt{3/5}\times\textsl{t}(5)-Y_i/2.7$. The covariance of $X_{i1},\dots,X_{i10}$ is the Toeplitz matrix with the first row $(1,0.8,0.8^2,\dots,0.8^9)$. $X_{i11},\dots,X_{i100}\stackrel{\text{\tiny i.i.d}}{\sim}  \textsl{N}(0,1)$, $i=1,\dots,10000$. We randomly select 10 labeled data from each class and the remaining are treated as unlabeled data for both training and tuning sets.

In addition to the $S^3$LDA and SELF methods for partially labeled data, the $\ell_1$-SVM and the sparse LDA method for both the labeled data only and the full data, we report the theoretical Bayes error as a baseline for comparison. We also show the performance of the oracle solution to $S^3$LDA and $\ell_1$-LDA on the solution path, which is obtained by selecting the optimal parameters based on the performance on the whole test data set. The oracle solution (denoted as `$(o)$' in tables and figures) can be viewed as the best possible result for a method on the whole solution path, given that the tuning parameter selection method can effectively find the true optimality.  It indicates the potential of a method.

\begin{table}[!t]\footnotesize
\centering
\begin{tabular}{c||c|c|c|c|c||c|c|c}
\hline
Data & $\ell_1$-SVM$_l$ & $\ell_1$-LDA$_l$ & SELF & $S^3$LDA     & $S^3$LDA(o) & $\ell_1$-SVM$_c$ & $\ell_1$-LDA$_c$ & Bayes \\\hline
Example 1 & 0.099       & 0.096			 & 0.096 &\textbf{0.094}& \textbf{0.084}             & 0.082            &0.082             & 0.080  \\
		      & (0.0020)    & (0.0019)   & (0.0015)     & (0.0021)     & (0.0008)          & (0.0003)         &(0.0002)          &   \\\hline
Example 2 & 0.088       & 0.080      & 0.109     & \textbf{0.075}       & \textbf{0.056}    &0.052             &0.065             & 0.033	\\
($s=1.3$)	 & (0.0033)   & (0.0032)   &(0.0024)  & (0.0034)     & (0.0018)          & (0.0030)         & (0.0007)         &   \\\hline
\end{tabular}
\caption{Simulation results for Examples 1 and 2: misclassification rate with standard error for 100 replications for $S^3$LDA, SELF, $\ell_1$-LDA and $\ell_1$-SVM. Subscript $l$ indicates the results for labeled data only. Subscript $c$ indicates the results for complete data with label information available. $S^3$LDA(o) indicates the oracle solution of $S^3$LDA. The Bayes error is also provided. The results for Example 1 show that $S^3$LDA has slightly better performance than $\ell_1$-LDA, $\ell_1$-SVM and SELF, with a greater potential (error for $S^3$LDA(o) $=0.084$ which is almost the complete data errors and the Bayes error). $S^3$LDA also performs well in an HDLSS setting such as Example 2. The table shows the result for $s=1.3$. The $S^3$LDA oracle solution is even better than the complete data $\ell_1$-LDA and close to that of the complete data  $\ell_1$-SVM. SELF does not perform well in this example, even worse than methods using the labeled data only, possibly due to the restrictions of graph-based methods.
}\label{tab:1}
\end{table}

Example 1 is a low-dimensional study, where the sample size of the labeled data is greater than the dimension. Example 2, 3 and 4 are based on the HDLSS setting. In Example 2, we fix the dimensionality and study the performances with the change of signal strength $s$. Example 3 and Example 4 demonstrate the performance of $S^3$LDA with the increase of dimensionality. We explore a Gaussian case in Example 3 and a non-Gaussian case in Example 4. 

For Examples 1 and 2, the misclassification rates of $S^3$LDA and other methods are shown in Table \ref{tab:1}. The numbers of false positives and false negatives are shown in Table \ref{tab:2}. A false positive occurs when a zero variable has a nonzero coefficient and a false negative occurs when a true variable has a zero coefficient value. As SELF is not a method designed for variable selection, we do not consider its false positives and false negatives.

The results for Example 1 in Table \ref{tab:1} show that $S^3$LDA has slightly better performance than the semi-supervised method (SELF) and the $\ell_1$-LDA and $\ell_1$-SVM (when they are applied to the labeled data), and that it has a great potential (error for the oracle solution of $S^3$LDA $=0.084$ which is almost the complete data errors and the Bayes error). In Table \ref{tab:2}, although $S^3$LDA is shown to have more false positives than the labeled data $\ell_1$-SVM and $\ell_1$-LDA, the complete data $\ell_1$-LDA is actually the worse (with 0.81 out of 1 in Example 1), suggesting that the problem could be due to the ineffectiveness of the LDA methods for variable selection in such low-dimensional data.

\begin{table}[!t]
	\centering
		\begin{tabular}{c|c|c|c|c|c||c|c}
		\hline
		\multicolumn{2}{c|}{Data}  & $\ell_1$-SVM$_l$  & $\ell_1$-LDA$_l$ & $S^3$LDA     & $S^3$LDA(o) & $\ell_1$-SVM$_c$ & $\ell_1$-LDA$_c$  \\\hline
		\multirow{4}{*}{Example 1}&\multirow{2}{*}{FP} & 0.09/1  & 0.28/1   & 0.64/1    & 0.45/1   & 0.03/1   & \underline{0.81}/1    \\
		          & & (0.024)          & (0.037)         & (0.039)     & (0.041)     & (0.015)         & (0.032)          \\\cline{2-8}
							&\multirow{2}{*}{FN}& 0/1     & 0/1              & 0/1          & 0/1          & 0/1      & 0/1                   \\
		          & & (0)               & (0)              & (0)          & (0)          & (0)              & (0)          \\\hline\hline
		\multirow{4}{*}{Example 2}&\multirow{2}{*}{FP}& 74.6/98  & 3.0/98 & 8.1/98   & 3.9/98  & 16.0/98   & \underline{63.5}/98    \\
		          & & (0.27)          & (0.12)         & (1.11)     & (0.16)     & (2.24)         & (0.46)            \\\cline{2-8}
							&\multirow{2}{*}{FN}& 0.48/2	     & 0.20/2     & 0.11/2       & 0.05/2      & 0/2        & 0/2            	  \\
		          & & (0.042)          & (0.034)         & (0.026)     & (0.018)     & (0)              & (0)            \\\hline
		\end{tabular}
		\caption{Simulation results for Examples 1 and 2: average number of false positives (FP) and false negatives (FN) with standard error for 100 replications. Although $S^3$LDA is shown to have more false positives than the labeled data $\ell_1$-SVM and $\ell_1$-LDA in Example 1, the complete data $\ell_1$-LDA is actually the worst. In Example 2, $S^3$LDA and $S^3$LDA(o) have much fewer false positives than both $\ell_1$-SVM methods and the complete data $\ell_1$-LDA, while they have fewer false negatives than both labeled-data-only methods.}
		\label{tab:2}
\end{table}

\begin{figure}[!t]\vspace{-1.5ex}\footnotesize
	\begin{center}
	\includegraphics[height=0.6\linewidth]{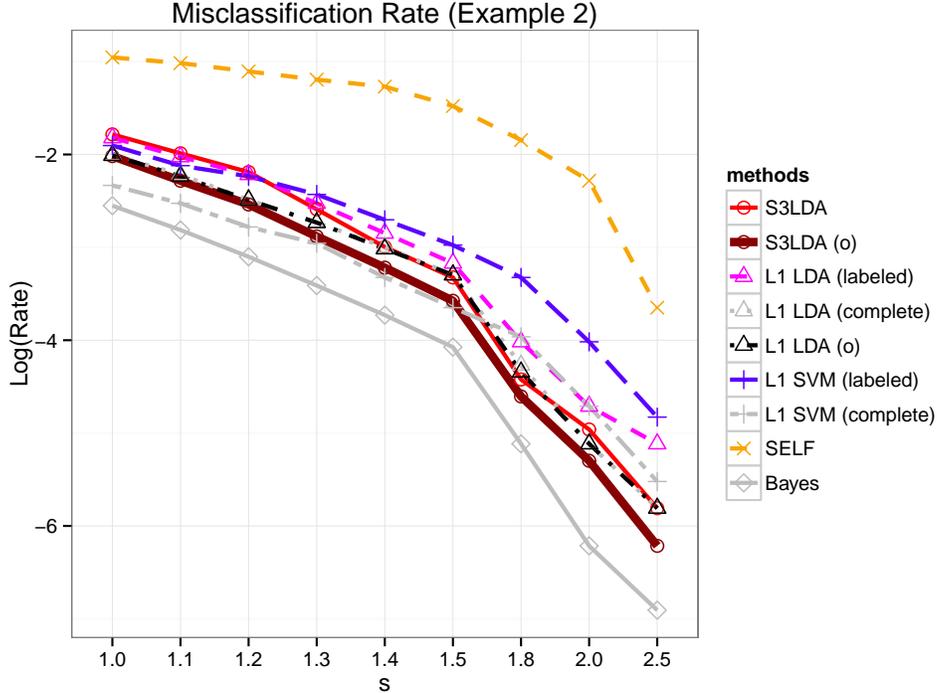}
	\end{center}\vspace{-2ex}
	\caption{Simulation results for Example 2 with various $s$ values. $S^3$LDA, $S^3$LDA(o) and the Bayes solutions are plotted in bold lines. It can be seen that $S^3$LDA (red line) does not perform well when $s$ is relatively small ($s=1.0,1.1,1.2$) but performs better than both labeled-data-only methods for larger $s$, and even better than both complete data methods for $s=1.8, 2.0, 2.5$. The oracle solution (dark red line) is always better than the complete data $\ell_1$-LDA and even its oracle version, which indicates a great potential of $S^3$LDA even when $s$ is relatively small. SELF does not perform well in this example possibly because the signal between the two classes is too small. As $s$ increases from 1.4 to 2.5, the misclassification rate of SELF dramatically decreases, which may suggest that it highly relies on the signal level.}
	\label{fig:example2_s}
\end{figure}

$S^3$LDA also performs well in the HDLSS setting such as Example 2. Table \ref{tab:1} and \ref{tab:2} show the results for the case $s=1.3$ in Example 2. $S^3$LDA again performs slightly better than the labeled data $\ell_1$-LDA and $\ell_1$-SVM, with the potential to be even better than the complete data $\ell_1$-LDA (error for $S^3$LDA oracle $=0.056$ which is less than the error for $\ell_1$-LDA$_c$ $=0.065$). This can possibly be explained by its variable selection performance, shown in the bottom half of Table \ref{tab:2}. $S^3$LDA and $S^3$LDA(o) have much fewer false positives than the $\ell_1$-SVM (for both complete data and for the labeled data only) and the complete data $\ell_1$-LDA, while they have fewer false negatives than both labeled-data-only methods. SELF again does not perform well in this example, even worse than methods using the labeled data only. This may be explained by the restriction of the graph-based methods such as SELF. Graph-based methods often reply on the adjacency matrix which describes the neighboring relationship between data points. Such a method encourages neighbors to be classified to the same class. Thus they are particularly useful when the signal is large, that is, two clusters can be easily identified. However, in Example 2, the two classes almost overlap with each other so that the adjacency matrix is not very helpful for classification. As the signal increases, SELF indeed gets better performance (see Figure \ref{fig:example2_s}).

We also study the performance of Example 2 with changing signal strength $s$, shown in Figure \ref{fig:example2_s}. We consider $s\in\{1.0,1.1,1.2,1.3,1.4,1.5,1.8,2.0,2.5\}$. Different lines and symbols represent different methods. $S^3$LDA, $S^3$LDA(o) and the Bayes solutions are plotted in bold lines. It can be seen that $S^3$LDA (red line) does not perform well when $s$ is relatively small ($s=1.0,1.1,1.2$) but performs better than both labeled-data-only methods for $s=1.3,1.4,1.5,1.8,2.0,2.5$, and even better than both complete data methods for $s=1.8, 2.0,2.5$. The oracle solution (dark red line) is always better than the complete data $\ell_1$-LDA and its oracle solution, which indicates a great potential of $S^3$LDA even when $s$ is relatively small. As discussed, SELF does not perform well in this example possibly because the signal between the two classes is too small. Figure \ref{fig:example2_s} shows that when $s$ increases from 1.4 to 2.5, the misclassification rate of SELF dramatically decreases, which suggests that it highly relies on the signal level.

\begin{figure}[!t]\vspace{-1.5ex}\footnotesize
  \begin{center}
  \includegraphics[height=0.5\linewidth]{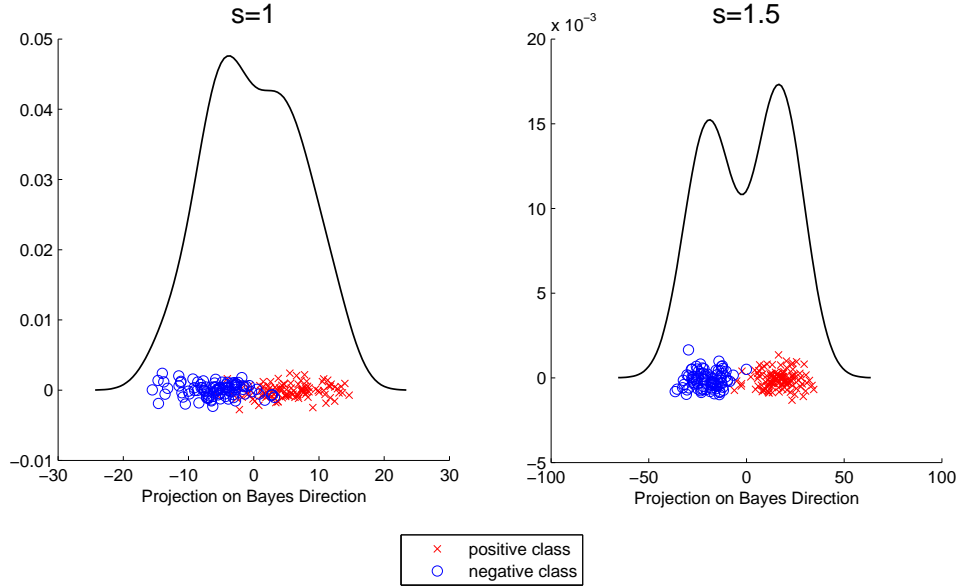}
  \end{center}\vspace{-2ex}
	\caption{The training data points in Example 2 projected onto the Bayes direction vector, for the cases $s=1$ (left) and $s=1.5$ (right). The positive and negative classes are represented in red and blue respectively. The $x$-axes show the projected values and the $y$-axes show random jitters for better visualization. Kernel density estimates are displayed as the curves.}
   \label{fig:example2}
 \end{figure}

To understand the reason $S^3$LDA does not work well for Example 2 when the signal is small but works well for larger $s$, we plot the training data points for $s=1$ and $s=1.5$, after projected onto the Bayes direction vector, in Figure~\ref{fig:example2}. In the right panel  ($s=1.5$), there is a valley in the density curve; in the left panel ($s=1$), even with the theoretically best Bayes direction, the gap between the two classes is invisible. For this reason, the unlabeled data fail to boost the classification performance for $S^3$LDA as no single coefficient direction can provide a wider margin than the others.

\begin{figure}[!t]\vspace{-1.5ex}\footnotesize
  \begin{center}
  \includegraphics[height=0.4\linewidth,width=0.9\linewidth]{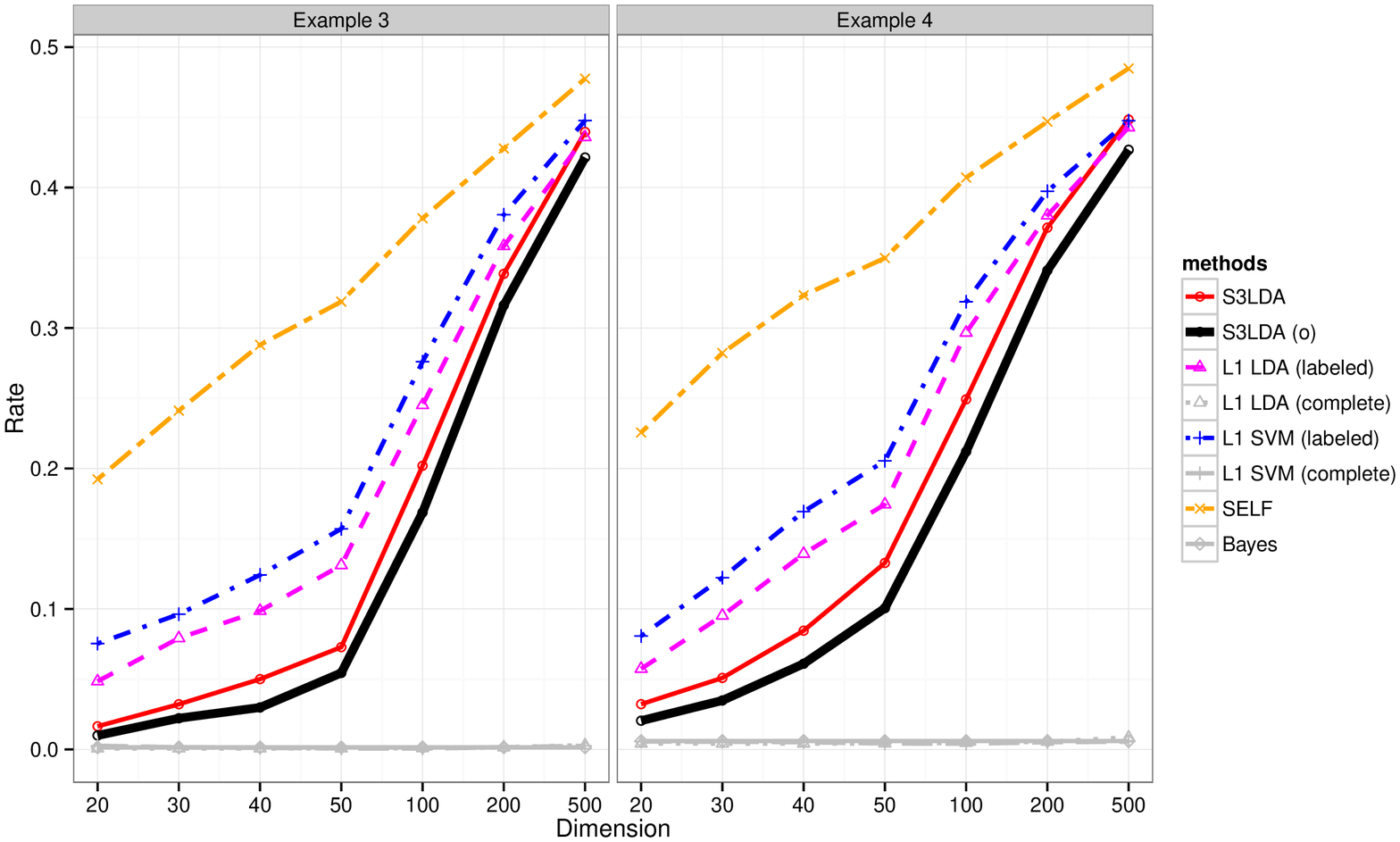}
   \vspace{-2ex}
  \end{center}\vspace{-1ex}
	\caption{Simulation results for Examples 3 and 4: misclassification rate over 100 replications for $S^3$LDA, $\ell_1$-LDA, $\ell_1$-SVM and SELF. The result for Example 3 is shown in the left panel and the result for Example 4 is shown in the right panel. For dimensions up to 100, $S^3$LDA (solid line) has a great improvement over $\ell_1$-LDA, $\ell_1$-SVM using labeled data alone, and SELF (dashed lines), with more potential (the oracle solution of $S^3$LDA, shown in bold line, is even better). $S^3$LDA does not provide much improvement for $d=200$ and $d=500$, in which cases the ratio of the dimensionality and the number of labeled data may be too large.}
   \label{fig:example3_4}
 \end{figure}

The misclassification rate results for Example 3 and 4 are displayed in Figure~\ref{fig:example3_4}. The result for Example 3 is shown on the left panel and the result for Example 4 is shown on the right panel. We explore a Gaussian case in Example 3 and a non-Gaussian case ($t$ distribution) in Example 4. Different methods are illustrated in different lines. The x-axis shows the dimensionality of the data. For dimensions up to 100, $S^3$LDA (solid line) has a great improvement, for both Gaussian and non-Gaussian cases, over $\ell_1$-LDA, $\ell_1$-SVM using labeled data alone, and SELF (dashed lines), with an even greater potential (the oracle solution of $S^3$LDA, shown in black bold line, is even better). SELF again fails to perform better than labeled data $\ell_1$-LDA and $\ell_1$-SVM in these two examples. The complete data $\ell_1$-SVM, complete data $\ell_1$-LDA and the Bayes error are also provided (all are close to zero). $S^3$LDA does not provide much improvement for $d=200$ and $d=500$, in which cases the ratio of the dimensionality and the number of labeled data may be too large.

\begin{figure}[!t]\vspace{-1.5ex}\footnotesize
  \begin{center}
  \includegraphics[height=0.48\linewidth]{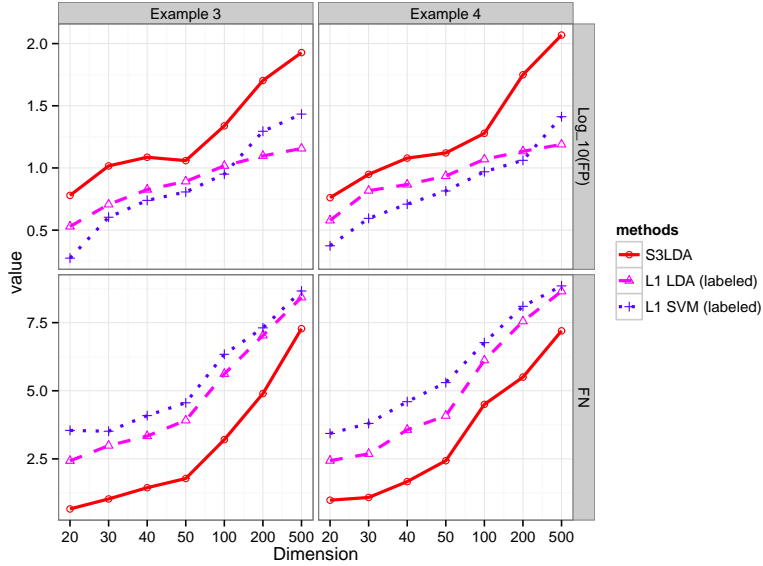}
   \vspace{-2ex}
  \end{center}\vspace{-1ex}
	\caption{Simulation results for Examples 3 and 4: the $\log_{10}$-false positives (top) and the false negatives (bottom) over 100 replications for $S^3$LDA (solid line), $\ell_1$-LDA (dashed line) and $\ell_1$-SVM (dotted line) using labeled data only. Both are shown with the increase of dimensionality. While $S^3$LDA performs poorly in terms of the false positives, it has fewer false negatives in both examples.}
   \label{fig:false_pos_neg}
 \end{figure}

In Figure~\ref{fig:false_pos_neg}, we compare the $\log_{10}$-false positives, and the false negatives for Example 3 and 4, over 100 replications, for $S^3$LDA, compared with the $\ell_1$-SVM and $\ell_1$-LDA for the labeled data only. The false positives, after taken the logarithm to the base 10, are shown in the top rows and the false negatives are shown in the bottom rows. While $S^3$LDA performs poorly regarding the false positives, it has fewer false negatives in both examples. 

\subsection{Real Data Application}
In this section, we analyze the Human Lung Carcinomas Microarray Data set using the $S^3$LDA method. This data set was previously analyzed in \cite{bhattacharjee2001classification}. Liu et al. \cite{liu2008statistical} used this data as a test bed to demonstrate their proposed significance analysis of clustering approach. The original data contain 12,625 genes. We have filtered the genes using the ratio of the sample standard deviation and sample mean of each gene and keep 2,530 of them with large ratios \cite{dudoit2002comparison, liu2008statistical}. We apply different methods to compare their classification performance.
The original Human Lung Carcinomas Data contains six classes. We combine the \textit{Squamous}, \textit{SmallCell} and \textit{Normal} subclasses to form the new positive class with sample size of 40, and combine the \textit{Colon} and \textit{Carcinoid} subclasses to form the new negative class with sample size of 37. Among the total of 77 observations, we randomly select 28 observations to keep the label information, 14 from each class, and treat the remaining 49 observations as unlabeled. The 28 labeled observations are split into 12 and 16 for training and tuning respectively. All the unlabeled data are used for training. We do not include the unlabeled data in the tuning criterion for this real data example, since the total amount of data is very limited and all the unlabeled data are used in training. We repeat the above procedure 100 times and report the average test errors on the unlabeled data in Table~\ref{realdata1}. 

\begin{table}[!ht]
	\centering
		\begin{tabular}{c|c|c|c|c||c|c}
		\hline
		          & $\ell_1$-SVM$_l$  & $\ell_1$-LDA$_l$ &SELF     & $S^3$LDA     & $\ell_1$-SVM$_c$ & $\ell_1$-LDA$_c$    \\\hline
		Error(\%) & 17.77             & 14.75            &16.77    & 7.49         & 0.31             & 0                  \\
		(SE(\%))  & (1.10)            & (0.91)           &(0.98)   &(0.74)       & (0.12)           & (0)          \\\hline
				
		\end{tabular}
		\caption{Averaged test errors as well as the estimated standard errors (in parenthesis) on a more complicated data over 100 independent replications of $S^3$LDA, SELF, $\ell_1$-penalized SVM and  $\ell_1$-penalized SVM with both labeled data only and complete data respectively.}
		\label{realdata1}
\end{table} 

In this challenging setting where both classes contain subclasses, $S^3$LDA works quite well, with $92.51\%$ classification accuracy. It outperforms labeled data $\ell_1$-SVM, labeled data $\ell_1$-LDA and SELF, with $12.50\%$, $8.83\%$ and $11.15\%$ improvements over the three methods respectively. Note that here the two classes are not as separated from each other, since both contain several possibly heterogeneous subclasses.

\section{Conclusion}

In this article, we propose a semi-supervised sparse Fisher's linear discriminant analysis method in the HDLSS setting. This method is designed for a dataset where a small amount of labeled data are available with a large amount of unlabeled data. In contrast to methods which rely on labeled data only, our method makes use of the unlabeled data to reconstruct the classification boundary. This is done by discouraging the boundary to go through an area with high density. 

For parameter tuning, we incorporate both the labeled and unlabeled data to identify the parameters with optimal performance. Our method outperforms $\ell_1$-LDA and $\ell_1$-SVM using labeled data alone in situations where the two classes have small overlap. Otherwise, $S^3$LDA performs as well as the $\ell_1$-LDA (for the labeled data only). In our numerical study, we often see great potentials of the $S^3$LDA through the competitive performance of the oracle solution. While in reality it is difficult to find the theoretically best tuning parameters with the very limited data, future research needs to be focused on improved parameter tuning criterion for the partially labeled data.

The $\ell_1$ norm penalty is used to handle the HDLSS data. Other penalty terms, such as elastic net, SCAD and MCP penalties, are possible as well. One possible reason that our $S^3$LDA does not perform well in terms of false positives is that the information contained in the zero variables in the unlabeled data has added additional noise that distracts the variable selection goal.

$S^3$LDA combines the classical model-based linear discriminant analysis and a machine learning oriented-technique. The LDA component fully uses the model assumptions and the covariance structure; the $U$ loss boosts the performance by extracting useful information from the unlabeled data. The proposed method has enriched the capacity of the LDA method and extend it to the partially labeled data territory. A related topic and future work is the significance analysis for partially labeled data \cite{Lu2015Significance}.

%
%

\section*{Acknowledgements}
Qiao's research is partially supported by Simons Foundation (award number 246649).  Both authors thank the Statistical and Applied Mathematical Sciences Institute (the 2012--2013 Program on Statistical and Computational Methodology for Massive Datasets) for their generous support where both authors have spent considerable amount of time when writing this article.

This material is based upon work partially supported by the National Science Foundation under Grant DMS-1127914 to the Statistical and Applied Mathematical Sciences Institute. Any opinions, findings, and conclusions or recommendations expressed in this material are those of the authors and do not necessarily reflect the views of the National Science Foundation.


\section*{Appendix}
\subsection*{Proof of Proposition~\ref{lemma1}}
Let the loss be $L(y,u)=(y-u)^2$. Since we assume the separating hyperplane goes through the origin, we consider 
\begin{align*}
Q(\omegav) & = \E[L(Y-\omegav'\Xv)]=\E[(Y-\omegav'\Xv)^2] \\
& =P(Y=+\delta)\E_{X|+\delta}[(Y-\omegav'\Xv)^2]+P(Y=-\delta)\E_{X|-\delta}[(Y-\omegav'\Xv)^2] \\
& =\half \int (\delta-\omegav'\xv)^2 \phi_+(\xv)d\xv+\half \int(-\delta-\omegav'\xv)^2\phi_-(\xv)d\xv,
\end{align*}
where $\phi_+(\xv)$ and $\phi_-(\xv)$ are the density functions of \textsl{N}$(\muv,\Sigma)$ and \textsl{N}$(-\muv,\Sigma)$ respectively.

The gradient of $Q(\omegav)$ is
\begin{align*}
\frac{\partial Q}{\partial\omegav} & = \int (\delta-\omegav'\xv)(-\xv) \phi_+(\xv)d\xv+\int(-\delta-\omegav'\xv)(-\xv)\phi_-(\xv)d\xv \\
& = -\delta\muv+\Sigma\omegav-\delta\muv+\Sigma\omegav \\
& = 2\Sigma\omegav-2\delta\muv.
\end{align*}
Then $\omegav_1=\delta\Sigma^{-1}\muv$.\qed

\subsection*{Proof of Proposition~\ref{lemma2}}
Let the loss be $L(u)=(1-|u|)_+$. Consider $Q(\omegav) = \E[L(\omegav'\Xv)]$. Note that without loss of generality, we assume that separating hyperplane goes through the origin. Our goal is to find 
\begin{align*}
	\omegav_2=\argmin_{\omegav:~\omegav'\muv=1}&~Q(\omegav)
\end{align*}
Hence, the Lagrangian is $Q(\omegav)-\alpha(\omegav'\muv-1)$.

We assume here $Y\in \{+\delta,-\delta\}$. Note $Q(\omegav) = \E[L(\omegav'\Xv)] = P(Y=+\delta)\E_{X|+\delta}[(1-|\omegav'\Xv|)_+]+P(Y=-\delta)\E_{X|-\delta}[(1-|\omegav'\Xv|)_+]$.

For $\E_{X|+\delta}[(1-|\omegav'\Xv|)_+]$, since we assume $\Xv\sim N_p(\muv,\Sigma)$, $\omegav'\Xv\sim N(\omegav'\muv,\omegav'\Sigma\omegav)$. Write $U := \omegav'\Xv$, and $Z:=\frac{U-\omegav'\muv}{\sqrt{\omegav'\Sigma\omegav}}$. We have 
\begin{align*}
	\E_{X|+\delta}[(1-|\omegav'\Xv|)_+]&=\E_U[(1-|U|)_+]=\int_{-1}^0(1+u)f_U(u)du+\int_{0}^1(1-u)f_U(u)du
\end{align*}
Here $\int_{-1}^0(1+u)f_U(u)du$ can be written, by change of variable, as 
\begin{align}\label{proofeq1}
	\int_{-1}^0(1+u)f_U(u)du = \int_{\frac{-1-\omegav'\muv}{\sqrt{\omegav'\Sigma\omegav}}}^{\frac{-\omegav'\muv}{\sqrt{\omegav'\Sigma\omegav}}}(1+\sqrt{\omegav'\Sigma\omegav}z+\omegav'\muv)f_Z(z)dz
\end{align}
Note that $(\sqrt{\omegav'\Sigma\omegav})'=\frac{\Sigma\omegav}{\sqrt{\omegav'\Sigma\omegav}}$ and $\left(\frac{\omegav'\muv}{\sqrt{\omegav'\Sigma\omegav}}\right)' = \frac{\sqrt{\omegav'\Sigma\omegav}\muv-\frac{\Sigma\omegav}{\sqrt{\omegav'\Sigma\omegav}}\omegav'\muv}{\omegav'\Sigma\omegav}=\frac{{\omegav'\Sigma\omegav}\muv-\Sigma\omegav\omegav'\muv}{(\omegav'\Sigma\omegav)^{3/2}}$.

The gradient of right hand side of (\ref{proofeq1}) is 
\begin{align*}
	\int_{\frac{-1-\omegav'\muv}{\sqrt{\omegav'\Sigma\omegav}}}^{\frac{-\omegav'\muv}{\sqrt{\omegav'\Sigma\omegav}}}(\frac{\Sigma\omegav}{\sqrt{\omegav'\Sigma\omegav}}z+\muv)f_Z(z)dz+1\cdot f_Z(\frac{-\omegav'\muv}{\sqrt{\omegav'\Sigma\omegav}})\cdot (-\frac{{\omegav'\Sigma\omegav}\muv-\Sigma\omegav\omegav'\muv}{(\omegav'\Sigma\omegav)^{3/2}})+0
\end{align*}

Similarly, $\int_{0}^1(1-u)f_U(u)du$ can be written, by change of variable, as 
\begin{align*}
	\int_{\frac{-\omegav'\muv}{\sqrt{\omegav'\Sigma\omegav}}}^{\frac{1-\omegav'\muv}{\sqrt{\omegav'\Sigma\omegav}}}(1-\sqrt{\omegav'\Sigma\omegav}z-\omegav'\muv)f_Z(z)dz
\end{align*}
whose gradient is
\begin{align*}
	\int_{\frac{-\omegav'\muv}{\sqrt{\omegav'\Sigma\omegav}}}^{\frac{1-\omegav'\muv}{\sqrt{\omegav'\Sigma\omegav}}}(-\frac{\Sigma\omegav}{\sqrt{\omegav'\Sigma\omegav}}z-\muv)f_Z(z)dz+0-1\cdot f_Z(\frac{-\omegav'\muv}{\sqrt{\omegav'\Sigma\omegav}})\cdot (-\frac{{\omegav'\Sigma\omegav}\muv-\Sigma\omegav\omegav'\muv}{(\omegav'\Sigma\omegav)^{3/2}})
\end{align*}

After taking a summation, we have that the gradient of $\E_{X|+\delta}[(1-|\omegav'\Xv|)_+]$ is 
\begin{align*}
	\int_{\frac{-1-\omegav'\muv}{\sqrt{\omegav'\Sigma\omegav}}}^{\frac{-\omegav'\muv}{\sqrt{\omegav'\Sigma\omegav}}}(\frac{\Sigma\omegav}{\sqrt{\omegav'\Sigma\omegav}}z+\muv)f_Z(z)dz+\int_{\frac{-\omegav'\muv}{\sqrt{\omegav'\Sigma\omegav}}}^{\frac{1-\omegav'\muv}{\sqrt{\omegav'\Sigma\omegav}}}(-\frac{\Sigma\omegav}{\sqrt{\omegav'\Sigma\omegav}}z-\muv)f_Z(z)dz
\end{align*}

The derivation for the gradient of $\E_{X|-\delta}[(1-|\omegav'\Xv|)_+]$ is similar, except that the mean of $X$ given $Y=-\delta$ is assumed to be $-\muv$. Hence its gradient is
\begin{align*}
	\int_{\frac{-1+\omegav'\muv}{\sqrt{\omegav'\Sigma\omegav}}}^{\frac{+\omegav'\muv}{\sqrt{\omegav'\Sigma\omegav}}}(+\frac{\Sigma\omegav}{\sqrt{\omegav'\Sigma\omegav}}z-\muv)f_Z(z)dz+\int_{\frac{\omegav'\muv}{\sqrt{\omegav'\Sigma\omegav}}}^{\frac{1+\omegav'\muv}{\sqrt{\omegav'\Sigma\omegav}}}(-\frac{\Sigma\omegav}{\sqrt{\omegav'\Sigma\omegav}}z+\muv)f_Z(z)dz
\end{align*}

The sum of the two gradients above, scaled by $P(Y=\delta)=P(Y=-\delta)=1/2$, is
\begin{align*}
	\frac{\partial Q_2(\omegav)}{\partial \omegav}=&\frac{\Sigma\omegav}{2\sqrt{\omegav'\Sigma\omegav}}\times A+\frac{\muv}{2}\Big\{\Phi(\frac{-\omegav'\muv}{\sqrt{\omegav'\Sigma\omegav}})-\Phi(\frac{-1-\omegav'\muv}{\sqrt{\omegav'\Sigma\omegav}})-\Phi(\frac{1-\omegav'\muv}{\sqrt{\omegav'\Sigma\omegav}})+\Phi(\frac{-\omegav'\muv}{\sqrt{\omegav'\Sigma\omegav}})\\
	&\quad -\Phi(\frac{+\omegav'\muv}{\sqrt{\omegav'\Sigma\omegav}})+\Phi(\frac{-1+\omegav'\muv}{\sqrt{\omegav'\Sigma\omegav}})+\Phi(\frac{1+\omegav'\muv}{\sqrt{\omegav'\Sigma\omegav}})-\Phi(\frac{+\omegav'\muv}{\sqrt{\omegav'\Sigma\omegav}})\Big\} \\
	=&b\Sigma\omegav+a\muv
\end{align*}

Define $G(m,n):=\int_{m}^n zf_Z(z)dz$. Note that $G(m,n) = -G(-n,-m)$. The term $A$ above can be found to be
\begin{align*}
	&G({\frac{-1-\omegav'\muv}{\sqrt{\omegav'\Sigma\omegav}}},{\frac{-\omegav'\muv}{\sqrt{\omegav'\Sigma\omegav}}})-G({\frac{-\omegav'\muv}{\sqrt{\omegav'\Sigma\omegav}}},{\frac{1-\omegav'\muv}{\sqrt{\omegav'\Sigma\omegav}}})\\
	&\quad\quad+G({\frac{-1+\omegav'\muv}{\sqrt{\omegav'\Sigma\omegav}}},{\frac{+\omegav'\muv}{\sqrt{\omegav'\Sigma\omegav}}})-G({\frac{\omegav'\muv}{\sqrt{\omegav'\Sigma\omegav}}},{\frac{1+\omegav'\muv}{\sqrt{\omegav'\Sigma\omegav}}})\\
	=&-G({\frac{\omegav'\muv}{\sqrt{\omegav'\Sigma\omegav}}},{\frac{1+\omegav'\muv}{\sqrt{\omegav'\Sigma\omegav}}})-G({\frac{-\omegav'\muv}{\sqrt{\omegav'\Sigma\omegav}}},{\frac{1-\omegav'\muv}{\sqrt{\omegav'\Sigma\omegav}}})\\
	&\quad\quad-G({\frac{-\omegav'\muv}{\sqrt{\omegav'\Sigma\omegav}}},{\frac{1-\omegav'\muv}{\sqrt{\omegav'\Sigma\omegav}}})-G({\frac{\omegav'\muv}{\sqrt{\omegav'\Sigma\omegav}}},{\frac{1+\omegav'\muv}{\sqrt{\omegav'\Sigma\omegav}}})\\
	=&-2\Big\{G({\frac{\omegav'\muv}{\sqrt{\omegav'\Sigma\omegav}}},{\frac{1+\omegav'\muv}{\sqrt{\omegav'\Sigma\omegav}}})+G({\frac{-\omegav'\muv}{\sqrt{\omegav'\Sigma\omegav}}},{\frac{1-\omegav'\muv}{\sqrt{\omegav'\Sigma\omegav}}})\Big\}
\end{align*}

Setting the gradient of the Lagrangian to be zero, we have $a\muv+b\Sigma\omegav=\alpha \muv$. Then $$\omegav_2\propto \Sigma^{-1}\muv.$$\qed

\subsection*{Proof of Theorem \ref{theorem1}}
Let $\delta=1/(\muv'\Sigma^{-1}\muv)$. Then in Proposition \ref{lemma1}, $\omegav_1=\delta\Sigma^{-1}\muv$; moreover, $\delta\Sigma^{-1}\muv$ satisfies the constraint $\omegav'\muv=1$ in Proposition \ref{lemma2}. Combining the results of Proposition \ref{lemma1} and Proposition \ref{lemma2}, we have $\omegav_{\infty}=\delta\Sigma^{-1}\muv$ for the joint optimization problem.\qed

\subsection*{Proof of Theorem \ref{theorem2}}
Let $\displaystyle{R(\omegav)=\E_{(\Xv,Y)}(Y-\omegav'\Xv)^2+C\E_{\Xv}(1-|\omegav'\Xv|)_+}$, $R_1(\omegav)=\E_{(\Xv,Y)}(Y-\omegav'\Xv)^2$ and $R_2(\omegav)=\E_{\Xv}(1-|\omegav'\Xv|)_+$, where $Y\in\{+\delta,-\delta\}$. For any $\omegav$, we have,
\begin{align*}
	R_1(\omegav)=&\E_{(\Xv,Y)}[(Y-\omegav'\Xv)^2]\\
	           =&\E_{(\Xv,Y)}[(Y-\omegav'\Xv)(Y-\omegav'\Xv)]\\
	           =&\E(Y^2)-2\E_{(\Xv,Y)}[(\omegav'\Xv)Y]+\E[\omegav'\Xv\Xv'\omegav]\\
	           =&\delta^2-2\delta\omegav'\muv+\omegav'\E[\E(\Xv\Xv'|Y)]\omegav \\
	           =&\delta^2-2\delta\omegav'\muv+\omegav'(\Sigma+\muv\muv')\omegav \\
	           =&\delta^2-2\delta\omegav'\muv+\omegav'\tilde\Sigma\omegav \\
	           \geq &\delta^2-2\delta\omegav'\muv+\lambda_{\min}(\tilde\Sigma)\|\omegav\|_2^2.
\end{align*}

Let $\omegav^{\lambda}=\omegav_{\infty}+\gammav^{\lambda}$. Recall that $\omegav_{\infty}=\delta\Sigma^{-1}\muv$. By the definition of $\omegav^{\lambda}$, we have
\begin{align*}
\gammav^{\lambda}&=\argmin_{\gammav:~(\omegav_{\infty}+\gammav)'\muv=1}R(\omegav_{\infty}+\gammav)+\lambda \|\omegav_{\infty}+\gammav\|_1 \\
								&=\argmin_{\gammav:~\gammav'\muv=0}\quad g(\gammav)+\lambda\sum_{k\in K^c}|\gamma_k|+\lambda\sum_{k\in K}(|\omega_{\infty}^k+\gamma_k|-|\omega_{\infty}^k|) \\
								&=\argmin_{\gammav:~\gammav'\muv=0}\quad f(\gammav)
\end{align*}
where $g(\gammav)=R(\omegav_{\infty}+\gammav)$ and $f(\gammav)$ is defined as $g(\gammav)+\lambda\sum_{k\in K^c}|\gamma_k|+\lambda\sum_{k\in K}(|\omega_{\infty}^k+\gamma_k|-|\omega_{\infty}^k|)$. We know that $f(\gammav^\lambda) - f(\0v)\le 0$ by the fact that $\gammav^\lambda$ minimizes $f$. Thus we have

\begin{align}\label{eq1}
g(\gammav^{\lambda})-g(\0v) &\leq \lambda\sum_{k\in K}(|\omega_{\infty}^k|-|\omega_{\infty}^k+\gamma^{\lambda}_k|)-\lambda\sum_{k\in K^c}|\gamma^{\lambda}_k|.
\end{align}

Note that $g(\gammav^{\lambda})-g(\0v)=[R_1(\omegav_{\infty}+\gammav^{\lambda})-R_1(\omegav_{\infty})] +C[R_2(\omegav_{\infty}+\gammav^{\lambda})-R_2(\omegav_{\infty})]$. For the first term $R_1(\omegav_{\infty}+\gammav^{\lambda})-R_1(\omegav_{\infty})$, we observe that
\begin{align}\label{eq2}
&R_1(\omegav_{\infty}+\gammav^{\lambda})-R_1(\omegav_{\infty}) \nonumber\\
= &\E_{(\Xv,Y)}[(Y-\omegav_{\infty}'\Xv-{\gammav^{\lambda}}'\Xv)^2]-\E_{(\Xv,Y)}[(Y-\omegav_{\infty}'\Xv)^2]\nonumber\\
= &\E_{(\Xv,Y)}[(Y-{\gammav^{\lambda}}'\Xv)^2-2\omegav_{\infty}'\Xv(Y-{\gammav^{\lambda}}'\Xv)+(\omegav_{\infty}'\Xv)^2]-\E_{(\Xv,Y)}[(Y-\omegav_{\infty}'\Xv)^2]\nonumber\\
= &R_1(\gammav^{\lambda})-2\delta\omegav_{\infty}'\muv+2\omegav_{\infty}'\tilde\Sigma\gammav^{\lambda}+\omegav_{\infty}'\tilde\Sigma\omegav_{\infty}-\E_{(\Xv,Y)}[(Y-\omegav_{\infty}'\Xv)^2]\nonumber\\
= &R_1(\gammav^{\lambda})-2\delta\omegav_{\infty}'\muv+2\omegav_{\infty}'\tilde\Sigma\gammav^{\lambda}+\omegav_{\infty}'\tilde\Sigma\omegav_{\infty}-[\delta^2-2\delta\omegav_{\infty}'\muv+\omegav_{\infty}'\tilde\Sigma\omegav_{\infty}]\nonumber\\
=&R_1(\gammav^{\lambda})+2\omegav_{\infty}'\tilde\Sigma\gammav^{\lambda}-\delta^2\nonumber\\
=&R_1(\gammav^{\lambda})+2\delta\muv'\Sigma^{-1}(\Sigma+\muv\muv')\gammav^{\lambda}-\delta^2\nonumber\\
=&R_1(\gammav^{\lambda})-\delta^2
\end{align}
Here the last statement is due to the constraint that $\gammav'\muv=0$.

For the second term $R_2(\omegav_{\infty}+\gammav^{\lambda})-R_2(\omegav_{\infty})$, we note that
\begin{align}\allowdisplaybreaks\label{eq3}
&R_2(\omegav_{\infty}+\gammav^{\lambda})-R_2(\omegav_{\infty})\nonumber\\
 =&\E_{\Xv}[(1-|\omegav_{\infty}'\Xv+{\gammav^{\lambda}}'\Xv|)_+]-\E_{\Xv}[(1-|\omegav_{\infty}'\Xv|)_+]\nonumber\\
						\geq& -\E_{\Xv}\Bigl|(1-|\omegav_{\infty}'\Xv+{\gammav^{\lambda}}'\Xv|)-(1-|\omegav_{\infty}'\Xv|)\Bigr|\nonumber\\
						=&-\E_{\Xv}\Bigl||\omegav_{\infty}'\Xv+{\gammav^{\lambda}}'\Xv|-|\omegav_{\infty}'\Xv|\Bigr|\nonumber\\
						\geq& -\E_{\Xv}\Bigl|\omegav_{\infty}'\Xv+{\gammav^{\lambda}}'\Xv-\omegav_{\infty}'\Xv\Bigr|\nonumber\\
						=&-\E_{\Xv}|{\gammav^{\lambda}}'\Xv|\nonumber\\
						\geq& -\sqrt{\E_{\Xv}[({\gammav^{\lambda}}'\Xv)^2]}\nonumber\\
						=&-\sqrt{{\gammav^{\lambda}}'\tilde\Sigma\gammav^{\lambda}}\nonumber\\
						\geq& -\sqrt{\lambda_{\max}(\tilde\Sigma)}\cdot\|\gammav^{\lambda}\|_2.
\end{align}
Combining (\ref{eq2}) and (\ref{eq3}), we have 
$$R_1(\gammav^{\lambda})-\delta^2-C\sqrt{\lambda_{\max}(\tilde\Sigma)}\cdot\|\gammav^{\lambda}\|_2\le g(\gammav^{\lambda})-g(\0v)$$
Due to this, along with (\ref{eq1}), we have
\begin{align*}
	R_1(\gammav^{\lambda})\le&\lambda\sum_{k\in K}(|\omega_{\infty}^k|-|\omega_{\infty}^k+\gamma^{\lambda}_k|)-\lambda\sum_{k\in K^c}|\gamma^{\lambda}_k|\\
	\quad& +\delta^2+C\sqrt{\lambda_{\max}(\tilde\Sigma)}\cdot\|\gammav^{\lambda}\|_2\\
	\le&\lambda\sum_{k\in K}(|\gamma^{\lambda}_k|)+\delta^2+C\sqrt{\lambda_{\max}(\tilde\Sigma)}\cdot\|\gammav^{\lambda}\|_2\\
	\le&\lambda\sqrt{s}\|\gammav^{\lambda}\|_2+\delta^2+C\sqrt{\lambda_{\max}(\tilde\Sigma)}\cdot\|\gammav^{\lambda}\|_2
\end{align*}
Recall that $$R_1(\gammav^{\lambda})\ge \delta^2-2\delta{\gammav^{\lambda}}'\muv+\lambda_{\min}(\tilde\Sigma)\|\gammav^{\lambda}\|_2^2=\delta^2+\lambda_{\min}(\tilde\Sigma)\|\gammav^{\lambda}\|_2^2$$

Combining the lower and upper bounds of $R_1(\gammav^{\lambda})$, we have
\begin{align*}
			&\delta^2+\lambda_{\min}(\tilde\Sigma)\|\gammav^{\lambda}\|_2^2	\le \lambda\sqrt{s}\|\gammav^{\lambda}\|_2+\delta^2+C\sqrt{\lambda_{\max}(\tilde\Sigma)}\cdot\|\gammav^{\lambda}\|_2\\
\Leftrightarrow\quad				&\lambda_{\min}(\tilde\Sigma)\|\gammav^{\lambda}\|_2^2\le \lambda\sqrt{s}\|\gammav^{\lambda}\|_2+C\sqrt{\lambda_{\max}(\tilde\Sigma)}\cdot\|\gammav^{\lambda}\|_2\\
\end{align*}
%
Thus 
		$$\|\gammav^{\lambda}\|_2\le \frac{\lambda\sqrt{s}+C\sqrt{\lambda_{\max}(\tilde\Sigma)}}{\lambda_{\min}(\tilde\Sigma)}.$$\qed

\bibliographystyle{ieeetr}
\bibliography{S3LDA_reference}
\end{document}